\documentclass[10pt, a4paper]{article}
\usepackage{lrec}
\usepackage{multibib}
\newcites{languageresource}{Language Resources}
\usepackage{graphicx}
\usepackage{tabularx}
\usepackage{soul}

\usepackage{epstopdf}
\usepackage[utf8]{inputenc}

\usepackage{hyperref}
\usepackage{xstring}

\usepackage[dvipsnames]{xcolor}
\usepackage{color}

\usepackage{amsmath}
\usepackage{amsfonts}
\usepackage{booktabs}

\title{Learning Word Vectors for 157 Languages}
\name{
  \hspace{-0.5em} Edouard Grave$^{1,*}$\thanks{$^*$ The two first authors contributed equally.} \hspace{0.2em} Piotr Bojanowski$^{1,*}$ \hspace{0.2em}
  Prakhar Gupta$^{1,2}$ \hspace{0.2em} Armand Joulin$^1$ \hspace{0.2em} Tomas Mikolov$^1$}
\address{
  $^1$Facebook AI Research \hspace{1em} $^2$EPFL \\
  \texttt{\{egrave,bojanowski,ajoulin,tmikolov\}@fb.com, prakhar.gupta@epfl.ch}
}

\abstract{
  Distributed word representations, or word vectors, have recently been applied to many tasks in natural language processing, leading to state-of-the-art performance.
  A key ingredient to the successful application of these representations is to train them on very large corpora, and use these pre-trained models in downstream tasks.
  In this paper, we describe how we trained such high quality word representations for 157 languages.
  We used two sources of data to train these models: the free online encyclopedia Wikipedia and data from the common crawl project.
  We also introduce three new word analogy datasets to evaluate these word vectors, for French, Hindi and Polish.
  Finally, we evaluate our pre-trained word vectors on 10 languages for which evaluation datasets exists, showing very strong performance compared to previous models.
  \\ \newline 
}

\begin{document}

\maketitleabstract

\section{Introduction}

Distributed word representations, also known as word vectors, have been widely used in natural language processing, leading to state of the art results for many tasks.
Publicly available models, which are pre-trained on large amounts of data, have become a standard tool for many NLP applications, but are mostly available for English.
While different techniques have been proposed to learn such representations~\cite{collobert2008unified,mikolov2013distributed,pennington2014glove}, 
all rely on the \emph{distributional hypothesis} -- the idea that the meaning of a word is captured by the contexts in which it appears.
Thus, the quality of word vectors directly depends on the amount and quality of data they were trained on.

A common source of data to learn word representations, available in many languages, is the online encyclopedia Wikipedia~\cite{al2013polyglot}.
This provides high quality data which is comparable across languages.
Unfortunately, for many languages, the size of Wikipedia is relatively small, and often not enough to learn high quality word vectors with wide coverage.
An alternative source of large scale text data is the web and resources such as the common crawl.
While they provide noisier data than Wikipedia articles, they come in larger amounts and with a broader coverage.

In this work, we contribute high quality word vectors trained on Wikipedia and the Common Crawl corpus, as well as three new word analogy datasets.
We collected training corpora for 157 languages, using Wikipedia and Common Crawl.
We describe in details the procedure for splitting the data by language and pre-processing it in Section~2.
Using this data, we trained word vectors using an extension of the fastText model with subword information~\cite{bojanowski2017enriching}, as described in Section~3.
In Section~4, we introduce three new word analogy datasets for French, Hindi and Polish and evaluate our word representations on word analogy tasks.
Overall, we evaluate our word vectors on 10 languages: Czech, German, Spanish, Finnish, French, Hindi, Italian, Polish, Portuguese and Chinese.
Our models for 157 languages other than English are available at \url{https://fasttext.cc}.

\paragraph{Related work.}
In previous work, word vectors pre-trained on large text corpora have been released alongside open source implementation of word embedding models.
English word vectors trained on a part of the Google News dataset (100B tokens) were published with \texttt{word2vec}~\cite{mikolov2013distributed}.
\newcite{pennington2014glove} released \texttt{GloVe} models trained on Wikipedia, Gigaword and Common Crawl (840B tokens).
A notable effort is the work of~\newcite{al2013polyglot}, in which word vectors have been trained for 100 languages using Wikipedia data.


\section{Training Data}
\label{sec:data}

We train our word vectors using datasets composed of a mixture of Wikipedia and Common Crawl.

\subsection{Wikipedia}
Wikipedia is the largest free online encyclopedia, available in more than 200 different languages.
Because the articles are curated, the corresponding text is of high quality, making Wikipedia a great resource for (multilingual) natural language processing.
It has been applied to many different tasks, such as information extraction~\cite{wu2010open}, or word sense disambiguation~\cite{mihalcea2007using}.
We downloaded the XML Wikipedia dumps from September 11, 2017.
The first preprocessing step is to extract the text content from the XML dumps.
For this purpose, we used a modified version of the \texttt{wikifil.pl} script\footnote{\url{http://mattmahoney.net/dc/textdata.html}} from Matt Mahoney.

Even if Wikipedia is available for more than 200 languages, many dumps are relatively small in size (compared to the English one).
As an example, some widely spoken languages such as Hindi, have relatively small Wikipedia data (39 millions tokens).
Overall, 28 languages contain more than 100 millions tokens, and 82 languages contain more than 10 millions tokens.
We give the number of tokens for the 10 largest Wikipedia in Table~\ref{tab:wikisize}.
For these reasons (and the fact that Wikipedia is restricted to encyclopedic domains), we decided to also use data from the common crawl to train our word vectors.

\begin{table}[t]
  \centering
  \begin{tabular}{lrr}
    \toprule
    Language & \# tokens & \# words \\
    \midrule
    German   & 1,384,170,636 & 3,005,294 \\
    French   & 1,107,636,871 & 1,668,310 \\
    Japanese &   998,774,138 &   916,262 \\ 
    Russian  &   823,849,081 & 2,230,231 \\
    Spanish  &   797,362,600 & 1,337,109 \\
    Italian  &   702,638,442 & 1,169,177 \\
    Polish     & 386,874,622 & 1,298,250 \\
    Portuguese & 386,107,589 &   815,284 \\
    Chinese    & 374,650,371 & 1,486,735 \\
    Czech      & 178,516,890 &   784,896 \\
    Finnish    & 127,176,620 &   880,713 \\
    Hindi      & 39,733,591 & 183,211 \\
    \bottomrule
  \end{tabular}
  \caption{
    Comparison of the size of the Wikipedia corpora for selected languages.
    The second column indicates the number of words which appear at least five times in the corpus.
  }
  \label{tab:wikisize}
\end{table}

\subsection{Common Crawl}
The common crawl is a non profit organization which crawls the web and makes the resulting data publicly available.
This large scale corpus was previously used to estimate $n$-gram language models~\cite{buck2014ngram} or to learn English word vectors~\cite{pennington2014glove}.
To the best of our knowledge, it was not used yet to learn word vectors for a large set of languages.
The data is distributed either as raw HTML pages, or as WET files which contain the extracted text data, converted to UTF-8.
We decided to use the extracted text data, as it is much smaller in size, and easier to process (no need to remove HTML).
We downloaded the May 2017 crawl, corresponding to roughly 24 terabytes of raw text data.

\begin{table}[t]
  \centering
  \setlength\tabcolsep{4pt} 
  \begin{tabular}{lcccccc}
    \toprule
    & \multicolumn{2}{c}{TCL} & \multicolumn{2}{c}{Wikipedia} & \multicolumn{2}{c}{EuroGov} \\
    \cmidrule(r){2-3} \cmidrule(r){4-5} \cmidrule(r){6-7}
    Model & Acc. & Time & Acc. & Time & Acc. & Time \\
    \midrule
    \texttt{langid.py} & 93.1 & 8.8 & 91.3 & 9.4 & 98.7 & 13.1 \\
    \texttt{fastText}  & 94.7 & 1.3 & 93.0 & 1.3 & 98.7 &  2.9 \\
    \bottomrule
  \end{tabular}
  \setlength\tabcolsep{6pt} 
  \caption{Accuracy and processing time of our language detector and \texttt{langid.py} on three publicly available datasets. The TCL dataset was converted to UTF-8.}
  \label{tab:langdect}
\end{table}

\paragraph{Language Identification.}
The first preprocessing step consists in splitting the data based on the language.
As noted by \newcite{buck2014ngram}, some pages contain text in different languages.
We thus decided to detect the language of each line independently.
For this purpose, we built a fast language detector using the \texttt{fastText} linear classifier~\cite{joulin2017bag}, which can recognize 176 languages.
We used 400 millions tokens from Wikipedia (described in the previous section) as well as sentences from the Tatoeba website\footnote{\url{www.tatoeba.org}} to train our language detector.
The model uses character ngrams of length 2, 3 and 4 as features, and a hierarchical softmax for efficiency.
We evaluate our model on publicly available datasets from \newcite{baldwin2010language} and report results in Table~\ref{tab:langdect}.
Our approach compares favorably to existing methods such as \texttt{langid.py}~\cite{lui2012langid}, while being much faster.
This language detector will be released along the other resources described in this article.
After language identification, we only keep lines of more than 100 characters and with a high confidence score ($\ge 0.8$).

\subsection{Deduplication and Tokenization}
The second step of our pipeline is to remove duplicate lines from the data.
We used a very simple method for this, computing the hash of each line, and removing lines with identical hashes (we used the default hash function of java String objects).
While this could potentially remove unique lines (which do not have a unique hash), we observed very little collision in practice (since each language is processed independently).
Removing duplicates is important for the crawl data, since it contains large amounts of boilerplate, as previously noted by \newcite{buck2014ngram}.
Overall, 37\% of the crawl data is removed by deduplication, while 21\% of the Wikipedia data is removed by this operation.

The final step of our preprocessing is to tokenize the raw data.
We used the Stanford word segmenter~\cite{chang2008optimizing} for Chinese, Mecab~\cite{kudo2005mecab} for Japanese and UETsegmenter~\cite{nguyen2016hybrid} for Vietnamese.
For languages written using the Latin, Cyrillic, Hebrew or Greek scripts, we used the tokenizer from the Europarl preprocessing tools~\cite{koehn2005europarl}. 
For the remaining languages, we used the ICU tokenizer.
We give statistics for the most common languages in Table~\ref{tab:wikisize} and \ref{tab:crawlsize}.

\begin{table}[t]
  \centering
  \begin{tabular}{lrr}
    \toprule
    Language & \# tokens & \# words \\
    \midrule
Russian    & 102,825,040,945 & 14,679,750 \\
Japanese   &  92,827,457,545 &  9,073,245 \\
Spanish    &  72,493,240,645 & 10,614,696 \\
French     &  68,358,270,953 & 12,488,607 \\
German     &  65,648,657,780 & 19,767,924 \\
Italian    &  36,237,951,419 & 10,404,913 \\
Portuguese &  35,841,247,814 &  8,370,569 \\
Chinese    &  30,176,342,544 & 17,599,492 \\
Polish     &  21,859,939,298 & 10,209,556 \\
Czech      &  13,070,585,221 &  8,694,576 \\
Finnish    &   6,059,887,126 &  9,782,381 \\
Hindi      &   1,885,189,625 &  1,876,665 \\
    \bottomrule
  \end{tabular}
  \caption{
    Comparison accross languages of the size of the datasets obtained using the Common Crawl.
    The second column indicates the vocabulary size of the models trained on this data.
  }
  \label{tab:crawlsize}
\end{table}


\section{Models}
\label{sec:model}

In this section, we briefly describe the two methods that we compare to train our word vectors.

\paragraph{Skipgram.}
The first model that we consider is the skipgram model with subword information, introduced by \newcite{bojanowski2017enriching}.
This model, available as part of the \texttt{fastText}\footnote{\url{https://fasttext.cc/}} software, is an extension of the skipgram model,
where word representations are augmented using character ngrams.
A vector representation is associated to each character ngram, and the vector representation of a word is obtained by taking the sum of the vectors of the character ngrams appearing in the word.
The full word is always included as part of the character ngrams, so that the model still learns one vector for each word.
We refer the reader to \newcite{bojanowski2017enriching} for a more thorough description of this model.

\paragraph{CBOW.}
The second model that we consider is an extension of the CBOW model~\cite{mikolov2013distributed}, with position weights and subword information.
Similar to the model described in the previous paragraph, this model represents words as bags of character ngrams.
The second difference with the original CBOW model is the addition of position dependent weights, in order to better capture positional information.
In the CBOW model, the objective is to predict a given word $w_0$ based on context words $w_{-n}, ..., w_{-1}, w_1, ..., w_{n}$.
A vector representation $\mathbf{h}$ of this context is obtained by averaging the corresponding word vectors:
$$
\mathbf{h} = \sum_{\substack{i=-n \\ i \neq 0}}^n \mathbf{u}_{w_i}
$$
Here, we propose to use the model with position weights introduced by \newcite{mnih2013learning}.
Before taking the sum, each word vector is multiplied (element wise) by a position dependent vector.
More formally, the vector representation $\mathbf{h}$ of the context is obtained using:
$$
\mathbf{h} = \sum_{\substack{i=-n \\ i \neq 0}}^n \mathbf{c}_i \odot \mathbf{u}_{w_i},
$$
where $\mathbf{c}_i$ are vectors corresponding to each position in the window, $\odot$ is the element-wise multiplication and $\mathbf{u}_{w_i}$ are the word vectors.
We remind the reader that the word vectors $\mathbf{u}_{w_i}$ are themselves sums over the character ngrams.
We refer the reader to \newcite{mikolov2017advances} for a study of the effect of deduplication and model variants (such as position-weighted CBOW)
on the quality of the word representations.


\section{Evaluations}
\label{sec:eval}

In this work, we evaluate our word vectors on the word analogy task.
Given a triplet of words \emph{A : B :: C}, the goal is to guess the word \emph{D} such that \emph{A : B} and \emph{C : D} share the same relation.
An example of such analogy question is \emph{Paris : France :: Berlin : ?}, where the corresponding answer is \emph{Germany}.
Word vectors can be evaluated at this task by computing the expected representation of the answer word \emph{D}.
Given word vectors $x_A$, $x_B$ and $x_C$ respectively for words \emph{A}, \emph{B} and \emph{C}, the answer vector can be computed as $x_B - x_A + x_C$.
In order to evaluate, the closest word vector $x$ in the dictionary is retrieved (omitting the vectors $x_A$, $x_B$ and $x_C$) and the corresponding word is returned.
Performance is measured using average accuracy over the whole corpus.

\subsection{Evaluation Datasets}
\label{sec:anal-data}

Analogy datasets are composed of word 4--uplets, of the form \emph{Paris : France :: Rome : Italy}.
Such datasets are usually composed of all the possible combinations of pairs such as \emph{Paris : France}, \emph{Berlin : Germany} or \emph{Beijing : China}.
In our evaluation, we use the dataset of~\newcite{svoboda2016new} for Czech, 
  that of~\newcite{koper2015multilingual} for German,
  that of~\newcite{cardellino2016spanish} for Spanish,
	that of~\newcite{venekoski2017finnish} for Finnish,
	that of~\newcite{berardi2015word} for Italian,
	the European variant of the dataset proposed by~\newcite{hartmann2017portuguese} for Portuguese and
	that of~\newcite{chen2015joint} for Chinese.

One of the contributions of this work is the introduction of word analogy datasets for French, Hindi and Polish.
To build these datasets, we use the English analogies introduced by~\newcite{mikolov2013efficient} as a starting point.
Most of the word pairs are directly translated, and we introduced some modifications, which are specific for each language.

\paragraph{French.} 
We directly translated all the word pairs in the \verb+capital-common-countries+, \verb+capital-world+ and \verb+currency+ analogies.
For \verb+family+ we translated most pairs, but got rid of ambiguous ones (singular and plural for \emph{fils}) or those that translate into nominal phrases.
We replaced the \verb+city-in-state+ category by capitals of French \emph{d\'epartements}, removing those where either the \emph{d\'epartement} or capital name is a phrase.
We also added a category named \verb+antonyms-adjectives+ composed of antinomic adjectives such as \emph{chaud} / \emph{froid} (hot and cold).
For syntactic analogies, we translated word pairs in all categories, except for \verb+comparative+ and \verb+superlative+, which in french are trivial: for example \emph{fort}, \emph{plus fort}, \emph{le plus fort} (strong, stronger, strongest).
When the word pair was ambiguous we either removed it or replaced with another one.
Finally, we added a new \verb+past-participle+ category with pairs such as \emph{pouvoir} and \emph{pu}.
In total, this dataset is composed of 31,688 questions.

\begin{table*}[t]
  \centering
  \begin{tabular}{l c cccccccccc c r}
    \toprule
    && \textsc{Cs} & \textsc{De} & \textsc{Es} & \textsc{Fi} & \textsc{Fr} & \textsc{Hi} & \textsc{It} & \textsc{Pl} & \textsc{Pt} & \textsc{Zh} && Average \\
    \midrule
    Baseline      && 63.1 & 61.0 & 57.4 & 35.9 & 64.2 & 10.6 & 56.3 & 53.4 & 54.0 & 60.2 && 51.0 \\
    $n$-gram 5-5  && 57.7 & 61.8 & 57.5 & 39.4 & 65.9 & 8.3  & 57.2 & 54.5 & 54.8 & 59.3 && 50.9 \\
    CBOW          && 63.9 & 71.7 & 64.4 & 42.8 & 71.6 & 14.1 & 66.2 & 56.0 & 60.6 & 51.5 && 55.5 \\
    +negatives    && 64.8 & 73.7 & 65.0 & 45.0 & 73.5 & 14.5 & 68.0 & 58.3 & 62.9 & 56.0 && 57.4 \\
    +epochs       && 64.6 & 73.9 & 67.1 & 46.8 & 74.9 & 16.1 & 69.3 & 58.2 & 64.7 & 60.6 && 58.8 \\
    \midrule
    Using Crawl   && 69.9 & 72.9 & 65.4 & 70.3 & 73.6 & 32.1 & 69.8 & 67.9 & 66.7 & 78.4 && 66.7 \\
    \bottomrule
  \end{tabular}
  \caption{
    Performance of the various word vectors on the word analogy tasks.
    We restrict the vocabulary for the analogy tasks to the 200,000 most frequent words from the training data.
  }
  \label{tab:anal}
\end{table*}

\begin{table*}[t]
  \centering
  \begin{tabular}{l c ccccccccccc }
    \toprule
    && \textsc{Cs} & \textsc{De} & \textsc{Es} & \textsc{Fi} & \textsc{Fr} & \textsc{Hi} & \textsc{It} & \textsc{Pl} & \textsc{Pt} & \textsc{Zh}  \\
    \midrule
    Wikipedia     && 76.9 & 79.1 & 93.9 & 94.6 & 88.1 & 70.8 & 80.9 & 69.5 & 79.2 & 100.0  \\
    Common Crawl  && 78.6 & 81.1 & 90.4 & 92.2 & 92.5 & 70.7 & 82.6 & 63.4 & 75.7 & 100.0  \\
    \bottomrule
  \end{tabular}
  \caption{
    Coverage of models trained on Wikipedia and Common Crawl on the word analogy tasks.
  }
  \label{tab:ana-cov}
\end{table*}

\paragraph{Hindi.}
All the word pairs in the categories \verb+capital-common-countries+, \verb+capital-world+ and \verb+currency+ were translated directly. 
For the \verb+family+ category, most of the pairs were translated. 
However, we got rid of word pairs like stepbrother and stepsister which translate into two-word phrases. 
Also, word-pairs which differentiate in the maternal or paternal origin of the relationship like `d\=ad\=a - d\=ad\=\i'  (paternal grandparents) and `n\=an\=a - n\=an\=\i' (maternal grandparents) were added. 
For the \verb+city-in-state+ category, city-state pairs from India were added, removing pairs in which the city or the state name is a phrase. 
We had to remove \verb+adjective-to-adverb+, \verb+comparative+, \verb+superlative+, \verb+present-participle+ and \verb+past-tense+ categories as in these cases, we are left with phrases rather than words. 
We also added a new category \verb+adjective-to-noun+, where an adjective is mapped to the corresponding abstract noun: for example `m\=\i\d{t}h\=a'(sweet)' is mapped to `mi\d{t}h\=as'(sweetness).

\paragraph{Polish.} 
As for the other languages, we translated all the word pairs in the \verb+capital-common-countries+, \verb+capital-world+, \verb+currency+ and \verb+family+ categories.
For the \verb+city-in-state+ category, we used the capital of Polish regions (\emph{wojew\'odztwo}).
For the syntactic analogies, we translated word pairs in all categories except for \verb+plural-verbs+, which we replaced with \verb+verb-aspect+.
One example with two aspects is \emph{iść} and \emph{chodzić} which are both imperfective verbs, but the second one expresses an aimless motion.
For the \verb+past-tense+ category, we use a mixture of perfective and imperfective aspects.
Overall, by taking all possible combinations, we come up with 24,570 analogies.

\subsection{Model Variants}
\label{sec:model-var}

In all our experiments, we compare our word vectors with the ones obtained by running the \texttt{fastText} skipgram model with default parameters -- we refer to this variant as ``Baseline''. 
Additionally, we perform an ablation study showing the importance of all design choices.
We successively add features as follows:
\begin{itemize}
  \item $n$-gram 5--5: getting word vectors with character $n$-grams of length 5 only.
    By default, the \texttt{fastText} library uses all character $n$-grams from length 3 to 6.
    One motivation for using fewer $n$-grams is that the corresponding models are much more efficient to learn.
  \item CBOW: using the model described in Sec.~\ref{sec:model} instead of the skipgram variant from~\newcite{bojanowski2017enriching}.
  \item +negatives: using more negative examples.
    By default, the \texttt{fastText} library samples 5 negative examples.
    Here, we propose to use 10 negatives.
  \item +epochs: using more epochs to train the models.
    By default, the \texttt{fastText} library trains models for 5 epochs.
    Here, we propose to train for 10 epochs.
  \item Using Crawl: instead of only training on Wikipedia, we also use the crawl data.
    For many languages, this corresponds to a large increase of training data size.
\end{itemize}

\subsection{Results}
\label{sec:anal}

We evaluate all the model variants on word analogies in ten languages and report the accuracy in Table~\ref{tab:anal}.
We restrict the vocabulary for the analogy tasks to the 200,000 most frequent words from the training data.
Therefore, the models trained on Wikipedia and Wikipedia+Crawl do not share the exact same vocabulary (see coverage in Table~\ref{tab:ana-cov}).

\paragraph{Influence of models and parameters.}
First, we observe that on average, all the modifications discussed in Section~\ref{sec:model-var} lead to improved accuracy on the word analogy tasks compared to the baseline \texttt{fastText} model.
First, using character $n$-grams of size 5, instead of using the default range of 3--6, does not significantly decrease the accuracy (except for Czech).
However, using a smaller number of character $n$-grams leads to faster training, especially when using the CBOW model.
Second, we note that using the CBOW model with position weights, described in Section~\ref{sec:model}, gives the biggest improvement overall.
Finally, using more negative examples and more epochs, while making the models slower to train, also leads to significant improvement in accuracy.

\paragraph{Influence of training data.}
One of the contributions of this work is to train word vectors in multiple languages on large scale noisy data from the web.
We now compare the quality of the obtained models to the ones trained on Wikipedia data.
Unsurprisingly, we observe that for high resources languages, such as German, Spanish or French, using the crawl data does not increase (or even slightly decreases) the accuracy.
This is partly explained by the domain of the analogy datasets, which corresponds well to Wikipedia.
However, it is important to keep in mind that the models trained on the crawl data have a larger coverage, and might have better performance on other domains.
Second, we observe that for languages with small Wikipedia, such as Finnish or Hindi, using the crawl data leads to great improvement in performance:
+23.5 for Finnish, +9.7 for Polish, +16.0 for Hindi, +17.8 for Chinese.

\section{Conclusion}
In this work, we contribute word vectors trained on Wikipedia and the Common Crawl, as well as three new analogy datasets to evaluate these models,
and a fast language identifier which can recognize 176 languages.
We study the effect of various hyper parameters on the performance of the trained models, showing how to obtain high quality word vectors.
We also show that using the common crawl data, while being noisy, can lead to models with larger coverage, and better models for languages with small Wikipedia.
Finally, we observe that for low resource languages, such as Hindi, the quality of the obtained word vectors is much lower than for other languages.
As future work, we would like to explore more techniques to improve the quality of models for such languages.

\section{Bibliographical References}
\label{main:ref}

\bibliographystyle{lrec}
\bibliography{xample}

\end{document}